\documentclass{SCIS2019}
\usepackage{graphicx}
\usepackage{color}
\usepackage{amsmath}
\usepackage{amssymb}
\usepackage{booktabs}
\usepackage{multirow}
\usepackage{textcomp}
\usepackage{hyperref}
\hypersetup{colorlinks,linkcolor=red,anchorcolor=blue,linktocpage=true,urlcolor=blue,citecolor=blue}
\usepackage{booktabs}
\newcommand{\tabincell}[2]{\begin{tabular}{@{}#1@{}}#2\end{tabular}}
\definecolor{mypurple}{RGB}{155, 48, 255}

\begin{document}
\ArticleType{RESEARCH PAPER}
\Year{2019}
\Month{}
\Vol{}
\No{}
\DOI{}
\ArtNo{}
\ReceiveDate{}
\ReviseDate{}
\AcceptDate{}
\OnlineDate{}

\title{Learning to Focus: Cascaded Feature Matching Network for Few-shot Image Recognition}{Title for citation}

\author[1]{Mengting Chen}{}
\author[1]{Xinggang Wang}{}
\author[2]{Heng Luo}{}
\author[2]{Yifeng Geng}{}
\author[1]{Wenyu Liu}{{liuwy@hust.edu.cn}}

\AuthorMark{Mengting Chen}

\AuthorCitation{Mengting Chen, Xinggang Wang, Heng Luo, et al}


\address[1]{Huazhong University of Science and Technology, Wuhan {\rm 430074}, China}
\address[2]{Horizon Robotics, Beijing {\rm 100080}, China}

\abstract{Deep networks can learn to accurately recognize objects of a category by training on a large number of annotated images. However, a meta-learning challenge known as a low-shot image recognition task comes when only a few images with annotations are available for learning a recognition model for one category. The objects in testing/query and training/support images are likely to be different in size, location, style, and so on. Our method, called Cascaded Feature Matching Network (CFMN), is proposed to solve this problem. We train the meta-learner to learn a more fine-grained and adaptive deep distance metric by focusing more on the features that have high correlations between compared images by the feature matching block which can align associated features together and naturally ignore those non-discriminative features. By applying the proposed feature matching block in different layers of the few-shot recognition network, multi-scale information among the compared images can be incorporated into the final cascaded matching feature, which boosts the recognition performance further and generalizes better by learning on relationships. The experiments for few-shot learning on two standard datasets, \emph{mini}ImageNet and Omniglot, have confirmed the effectiveness of our method. Besides, the multi-label few-shot task is first studied on a new data split of COCO which further shows the superiority of the proposed feature matching network when performing few-shot learning in complex images. The code will be made publicly available.}


\maketitle

\section{Introduction}
Deep learning achieves great success in a variety of tasks with large amounts of labeled data for image recognition \cite{he2016deep,krizhevsky2012imagenet,simonyan2014very}, machine translation \cite{wu2016google,bahdanau2014neural} and speech synthesis \cite{oord2016wavenet}. However, labeled data is not always massively available when annotation cost is too expensive or time is not allowed. By contrast, the human can learn novel concepts with only a few examples in a short time \cite{bloom2000children}.

Few-shot learning attempts to resolve this problem by training a model that classifies an unlabeled example based on a small labeled support set. Specifically, $N$-way $K$-shot learning is the task of classifying an example, termed as a query, into one of $N$ classes, when only $K$ samples per class are available as supervision; these $N\times K$ samples with labels are termed as a support set. During training, support images and some query images are sampled. The meta-learner needs to distinguish the category of query images using only the support images. Moreover, the categories of the training set disjoint with those of testing set and they are randomly sampled to prevent direct semantic relationship and visual similarities between. Referring to \cite{vinyals2016matching}, the batch of the support set and queries is termed as an \textit{episode}.

\begin{figure}
  \centering
  \includegraphics[width=0.8\linewidth]{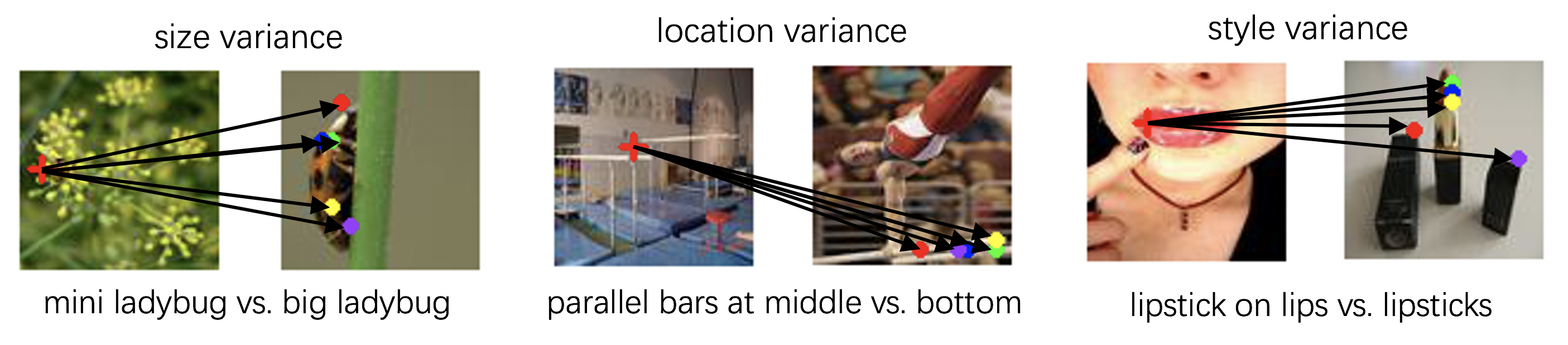}
  \caption{{\color{black}\textbf{Visualization of the feature matching results of CFMN.} Two adjacent images form a group. The feature at the red cross in the left query image matches with all features at the colored positions in the right support image. The colors, in turn, the red, green, blue, yellow, and purple point the positions which have the top five highest correlation responses. Although the interested objects may be different in size, location, style, and so on, they are associated together by our feature matching operation. More examples are shown in Fig.~\ref{visualization}. }}
  \label{introduction}
\end{figure}

Given a test image, the model of few-shot learning needs to estimate the feature similarities between the test and the supporting images of each class. Different from the traditional image recognition task that each class is represented by a parametric model learned from a large number of images, the category is supported by only a few, even a single image in the few-shot setting. This means that the classifier needs to accurately evaluate the similarity with a little supervision and strong variance due to the lack of enough supporting information. As can be seen from Fig.~\ref{introduction}, the query image may share limited visual similarities with support images. However, it's hard for the model to generalize among those strong inter-class differences with limited number of training images per class. \cite{pmlr-v70-finn17a, snell2017prototypical} show few-shot learning problem is prone to severe overfitting. To deal with the strong inter-class difference, we propose that the meta-learner should focus on essential spatial relationship features that have correlations between the query and support images and pay less attention to the non-discriminative features. We design our feature matching block to align the features of two compared images by the similarity of every feature position pairs. As shown in Fig.~\ref{introduction}, two positions corresponding to the object from the same category will get a high response by our method, even the overall images may look quite different visually. 

To fully utilize the proposed feature matching block, we apply three blocks at different layers of the network and cascade them together. The representation level of features from shallow layers of CNN is different from and usually lower than those of deep layers, and we extract the relation and similarity information of edges, shapes, and colors using shallow layers while deeper layers can produce object parts or other semantic information. The cascaded structure fuses all the information to make the final decision more accurate and robust.

In this paper, our main contributions are reflected in four aspects. \textbf{(1)} We propose a feature matching block that is capable of associating the object parts with high correlations between compared images and encouraging the model to pay more attention to those parts, which generalize to the large intra-class variation between the query and support images for few-shot learning challenge. \textbf{(2)} We cascade the feature matching block to obtain multi-scale representation. The cascaded structure obtains more robust and meaningful features (as can be shown in Fig.~\ref{introduction}) for the few-shot image recognition task. \textbf{(3)} The multi-label few-shot classification task is first proposed in this paper which shows the effectiveness of the proposed method for few-shot learning from a more realistic and complex sample space. A new split of COCO termed as FS-COCO, is compiled to benchmark this difficult yet important few-shot learning task. \textbf{(4)} We also evaluate the cascaded architecture model on Omniglot and \emph{mini}ImageNet. Our model shows state-of-the-art results. \textbf{(5)} We construct four hard settings of Omniglot to evaluate the model's robustness on size, location, and rotation variations.

\section{Related work} \label{Related_work}

\subsection{Deep learning for few-shot image recognition}

Few-shot image recognition is a challenging problem which gains increasing attention in recent years. A lot of deep learning techniques have sprung up. In order to increase memory capacity, some works adopted Neural Turing Machines \cite{graves2014neural, santoro2016meta} or LSTM \cite{hochreiter1997long,pmlr-v70-munkhdalai17a}. There are also some works using parameter adaptation. In MAML \cite{pmlr-v70-finn17a}, the parameters are explicitly trained to generalize well on new tasks by a small number of gradient steps with a small amount of training data. Sachin and Hugo \cite{ravi2016optimization} propose an LSTM based meta-learner model to learn the exact optimization algorithm used to train another learner neural network classifier in the few-shot regime. 

There are also some specialized neural networks for few-shot image recognition. Matching Network \cite{vinyals2016matching} learns an embedding function with a sample-wise attention kernel to predict the similarity. Compared to Matching Network, Prototypical Network \cite{snell2017prototypical} has a similar structure but employs Euclidean distance instead of cosine distance. {\color{black}TADAM \cite{NIPS2018_7352} proposes a dynamic task conditioned feature extractor based on Prototypical Network. Different from simple metrics,} Relation Network \cite{sung2017learning} learns a deep non-linear distance metric for similarity comparing. There are also some methods learn to predict the parameters for novel categories without additional training~\cite{qiao2017few, qi2018low, gidaris2018dynamic, oreshkin2018tadam}, learn as a regression problem \cite{bertinetto2018meta}, learn from unlabeled data~\cite{wang2016learning} or weakly-labeled data~\cite{liu2019prototype}. SNAIL uses temporal convolutions and soft attention to combine with the context of support samples. TPN \cite{liu2018learning} performs transductive learning on the similarity graph. DTN \cite{chen2020diversity} generates new reference features by transferring diversity information between training image pairs in the same class. \cite{weijian2021constellation} learns object parts by clustering cell features and modeling their relationships in an attentional manner for few shot learning, which obtains the state-of-the-art performance on the few-shot image classification benchmarks.

Data augmentation using generative models is also an effective option for few-shot learning \cite{Zhu_2017_ICCV,goodfellow2014generative}. At first, attributed-guided augmentation methods in feature space are used in AGA  \cite{Dixit_2017_CVPR} and FATTEN \cite{Liu_2018_CVPR}. Then Hariharan and Girshick \cite{Hariharan_2017_ICCV} transfer the transformation from a pair of known samples to a sample from a novel class. $\Delta$-encoder \cite{NIPS2018_7549} has similar target as \cite{Hariharan_2017_ICCV}, but it is trained as a reconstruction task. \cite{Wang_2018_CVPR} is more straightforward which generates samples by adding random noises to support features. There are also some methods used extra information, such as a deformation sub-network \cite{chen2019deformation} or a pre-trained saliency network \cite{Zhang_2019_CVPR}.

Our work is a specialized neural network that can establish semantic associations between images and encourage the model to focus more on the features that have high correlations; it overcomes the variance of inter-class and gets better performance for few-shot image recognition. 

\subsection{Matching and attention for few-shot image recognition}

Matching is an effective way to establish semantic correspondences between images \cite{thewlis2016fully,novotny2017anchornet,Wang_2018_CVPR}; and the attention mechanism can help to decide which features are more useful based on the established correspondences \cite{bahdanau2014neural,xu2015show,yang2016stacked}. Matching Network \cite{vinyals2016matching} uses the softmax function over the cosine distance between embedding features as a sample-wise attention kernel. It treats each image as an individual sample without differentiating the semantic meanings of different pixels. In our work, the attention is feature-wise between the query with each support images. It can learn the semantic correspondences between each feature pair in different positions. 

Attention can also be applied between label semantics and image domains \cite{wang2017multi, chu2018learning} for few-shot image recognition, but they need extra information for word embedding. Our method learns from the training images only, without any other external information. Our attention mechanism is similar to self-attention \cite{cheng2016long,parikh2016decomposable} which has proven to be effective on machine translation \cite{vaswani2017attention}, image transformer \cite{parmar2018image}, video sequence \cite{wang2017non} and GAN \cite{zhang2018self}. Self-attention aims to find the relations within an image/sequence, but our method focuses more on establishing the correlation responses of each feature position between images for more accurate similarity measure which is specially designed for few-shot image recognition. The STANet~\cite{yan2019dual} is also similar to us. But we combine the attention results from different feature expression levels, while STANet only uses the high-level feature. DCN \cite{zhang2018deep} is also based on the Siamese structure to learn the relation between the query and support image. A sequence of relation modules is used to compute a non-linear metric. But our cascaded matching block focuses on matching fine-grained similarity of two compared images, and highlights the corresponding feature to avoid interference from intra-class variance.

\section{Method} \label{Method}

\subsection{Problem definition} \label{PD}

To illustrate the few-shot image recognition task, we follow the definition in \cite{vinyals2016matching} which is termed as $N$-way $K$-shot learning. Each evaluation step is an $N$-way $K$-shot task which consists of two parts, support set, and query. We first sample $N$ classes from the training/testing set, then sample a support set $\mathcal{D}_s=\{(x_s^i,y_s^i), i\in [1,...,N\times K]\}$, which contains $K$ labeled examples from each of the $N$ classes. The query image $(x_q, y_q)$ is sampled from the rest images of the $N$ classes, \textit{i.e.} $y_q \in \{y_s^i, i\in [1,..., N\times K]\}$ and $x_q \notin \{x_s^i, i\in [1,..., N\times K]\}$. It needs to be classified into one of the $N$ classes based only on the support set. Different from traditional image recognition tasks based on lots of training images, the label space of the training set here is disjointed with it of the testing set. The testing process is in the form of $N$-way $K$-shot but with classes unknown to the training set.

\begin{figure}[!t]
  \centering
  \includegraphics[width=1\linewidth]{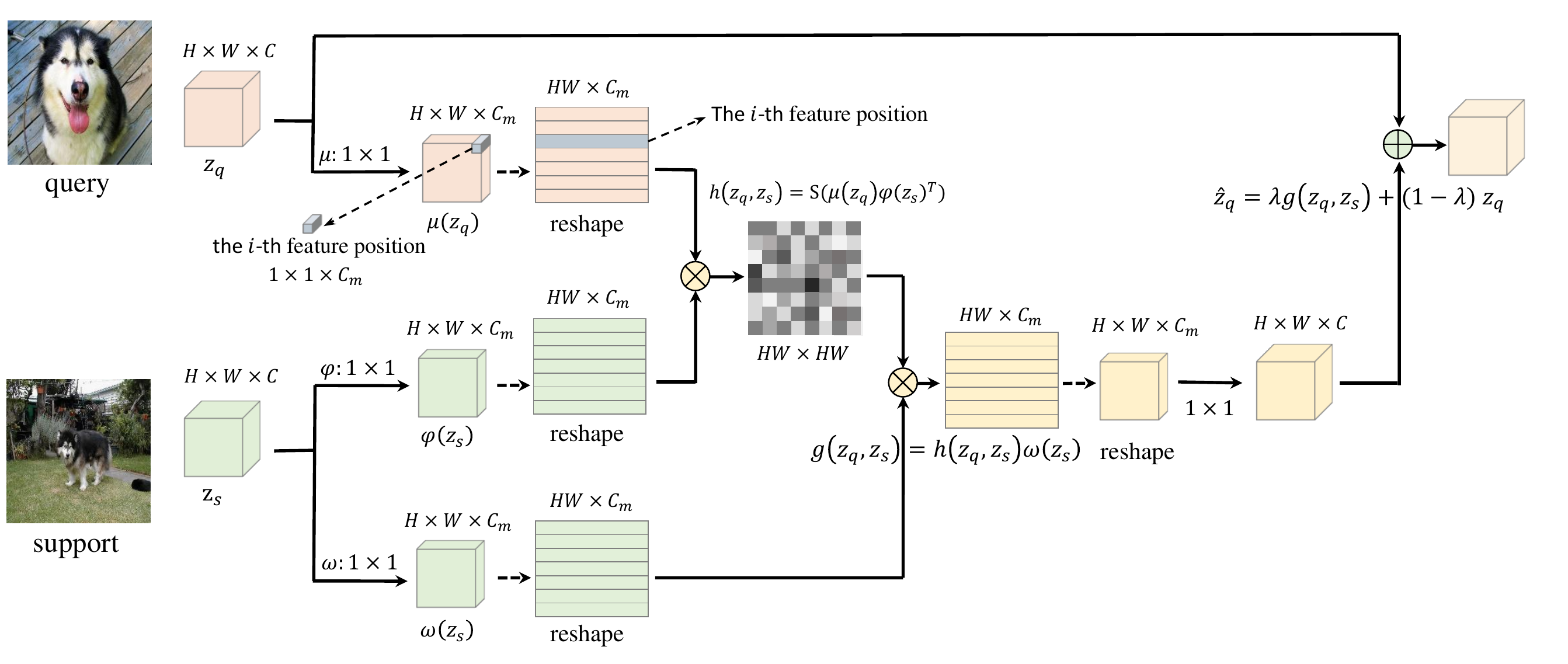}
  \caption{\textbf{Feature matching block.} $z_q$ and $z_s$ are the features of the query and support image, respectively, which have the same shape $H\times W\times C$. After the space transformation $\mu$, $\varphi$ and the reshape operation, $h(z_q,z_s)=S(\mu(z_q)\varphi(z_s)^T)$ is a spatial attention map between each feature position of the query and it of the support image. $S$ is the row-wise softmax. The feature $\omega(z_s)$ is scaled by the spatial attention map and mapped back to the input space. The final output of the block is the combination of the matched feature $g(z_q,z_s)$ and the original query feature $z_q$ with the proportion of $\lambda:(1-\lambda)$.}
  \label{nonlocal_1}
\end{figure}

\subsection{Feature matching block} \label{FMB}

The details of the feature matching block are shown in Fig.~\ref{nonlocal_1}. $z_q$ and $z_s$ are the features of the query and a support image from one of the hidden layers respectively, which are both in the shape of $H\times W\times C$. Firstly, they are mapped into another space $\mu$ and $\varphi$ to get $\mu(z_q)$ and $\varphi(z_s)$ respectively. Then they are reshaped to 2-dimensional matrices with the shape of $HW\times C_m$. The two matrices calculate a spatial attention map as follows,
\begin{equation} \label{eq:1}
h(z_q,z_s)=S(\mu(z_q)\varphi(z_s)^T),
\end{equation}
\noindent
where $S$ is the row-wise softmax. In the 3-dimensional metrics $\mu(z_q)$ and $\varphi(z_s)$, each feature point in $H\times W$ dimension is a feature position with the shape of $1\times 1\times C_m$, represented by $\mu(z_q^i)$ and $\varphi(z_s^i)$, $i\in[1,2,...,H\times W]$. After reshaping, each row of the 2-dimensional matrix is a feature position which is shown in In Fig.~\ref{nonlocal_1}. Therefore each element $h^{i,j}$ of the spatial attention map is the similarity between the feature in the $i$-th position of the query and the feature in the $j$-th position of the support image as defined as follows, 
\begin{equation} \label{eq:2}
h^{i,j} = \frac{exp(\mu(z_q^i)\varphi(z_s^j)^T)}{\sum _{j=1}^{H\times W}exp(\mu(z_q^i)\varphi(z_s^j)^T)}.
\end{equation}

Meanwhile, the support feature $z_s$ is mapped to another space $\omega$. It is scaled by the spatial attention map $h(z_q,z_s)$ to get $g(z_q,z_s)=h(z_q,z_s)\omega(z_s)$. Therefore, $\omega(z_s^j)$ indicates the feature in the $j$-th position of $\omega(z_s)$. A single feature position in $g(z_q, z_s)$ can be represented as follows, 
\begin{equation} \label{eq:3}
g^i = \sum _{j=1}^{H\times W}{h^{i,j}\omega(z_s^j)}.
\end{equation}

We can find that the $i$-th feature position of the feature map $g(z_q,z_s)$ depends on the correlation responses between the $i$-th feature position of the query $\mu(z_q)$ with all the feature positions of the support $\varphi(z_s)$. That is why we term it as a spatial attention mechanism. The features of $z_q$ and $z_s$ will be more retained if they are highly relevant to each other and the irrelevant features tend to be ignored. Then the network can learn to focus more on the relevant features, thereby reducing the influence of strong variance and producing better results. Then the matched feature $g(z_q,z_s)$ is mapped via a $1\times 1$ convolution layer to get the same shape as the input $z_q$ and $z_s$. Moreover, we find that keeping the original feature of the query image is helpful. In few-shot image recognition, in order to reach better similarity measurement, not only should the model focus on some particular parts that have high correlation responses, but also takes the whole feature into account. So the final output of the feature matching block is the combination of the matched feature $g(z_q,z_s)$ and the original query feature $z_q$ with the proportion of $\lambda:(1-\lambda)$ is described as follows,
\begin{equation} \label{eq:4}
\hat{z_q} = \lambda g(z_q,z_s)+(1-\lambda) z_q,
\end{equation}
where $\lambda$ is a weight factor over the matched feature. No matching information is injected if $\lambda=0$; only the matched features are considered if $\lambda=1$. 

\begin{figure*}
  \centering
  \includegraphics[width=1\linewidth]{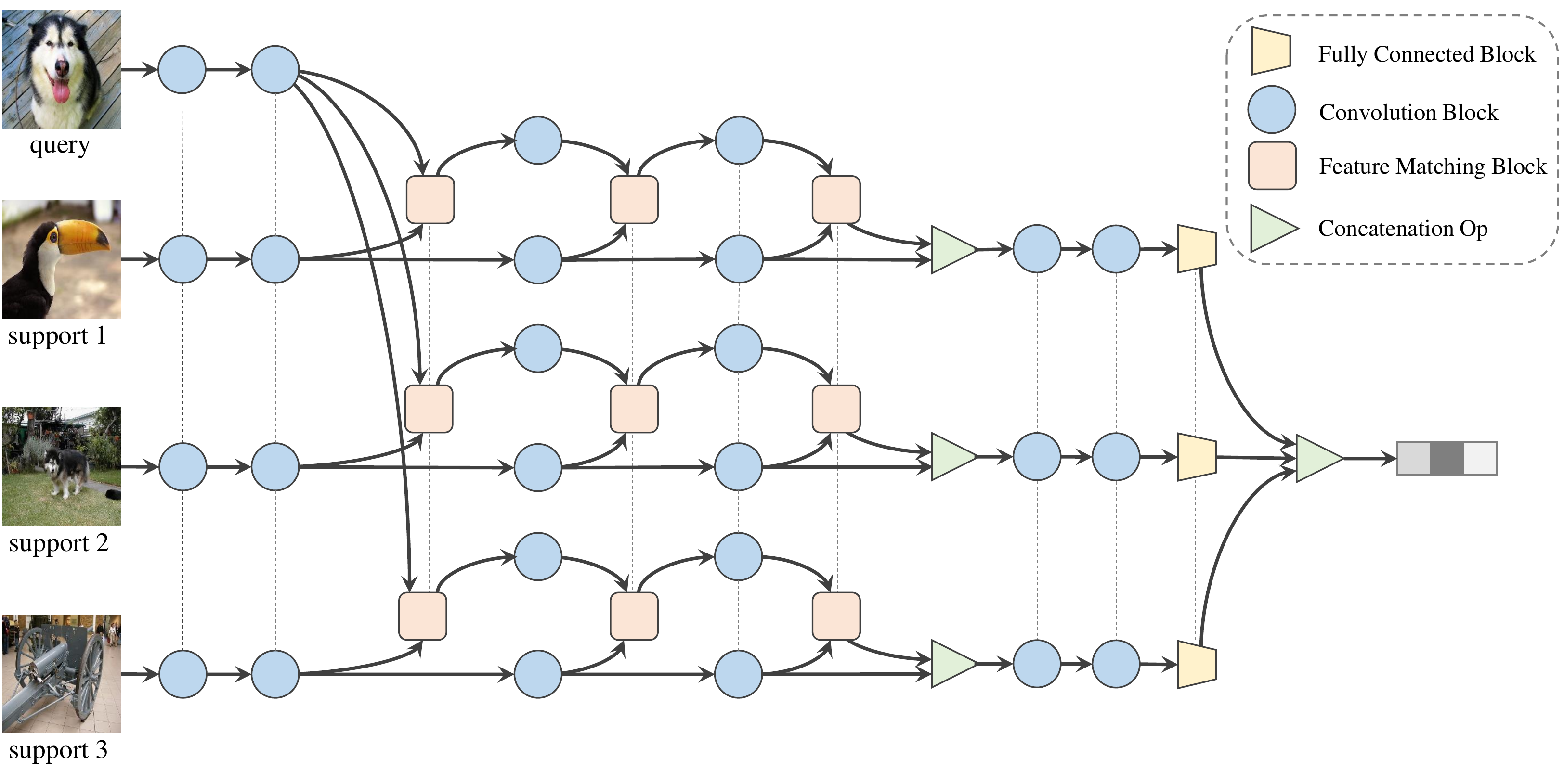}
\caption{\textbf{Illustration of the proposed Cascaded Feature Matching Network.} As shown in the top right corner of the figure, there are three different network blocks and one operation in CFMN. The blocks connected by a dashed line share the same parameters. Before the first Concatenation Operation, there are four Convolutional Blocks to extract the feature of each image. Three Feature Matching Blocks are applied after the second, the third, and the fourth Convolutional Blocks which form a cascaded structure. There are two Convolutional Blocks and one Fully Connected Block to predict the similarity of the two concatenated features. The final prediction is the connection of all the similarity scores}
  \label{nonlocal_2}
\end{figure*}

\subsection{Cascaded Feature Matching Network} \label{CFMN}

In Fig.~\ref{nonlocal_2}, we take $3$-way $1$-shot for example. The overall structure is a conditional neural network $f(x_q,D_s;\theta)$ as we described in Sect.~\ref{PD}. The input consists of the query $x_q$ (test image) and the support set $D_s$ (condition). The output of the network is a 3-dimensional vector which indicates the prediction for $x_q$. The class with the highest prediction value is the final categorized result.

The first four Convolutional Blocks and all the three Feature Matching Blocks can be viewed as a feature extractor. However, the extraction process of the query image is dependent on the feature matching results with respective support images. The cascaded structure combines matched information from different representation levels to reach a more accurate and robust performance.

After the feature extraction process, extracted features of the query and support images are concatenated in the channel dimension. Two Convolution Blocks and the Fully Connected Block after the first Concatenation Operation learn a distance metric of the concatenated feature. The output of the Fully Connected Block is a single value in a range of $[0, 1]$. The final output is the concatenation of all the three outputs of the Fully Connected Block. 

For $K$-shot where $K>1$, the query will get $K$ concatenated features with all $K$ support images for one class. We element-wise average over those $K$ concatenated features to predict one similarity score for this class. Thus, it can be guaranteed that there are only $N$ scores to form the final output.

\subsection{CFMN for multi-label few-shot classification}

\begin{figure}
  \centering
  \includegraphics[width=0.55\linewidth]{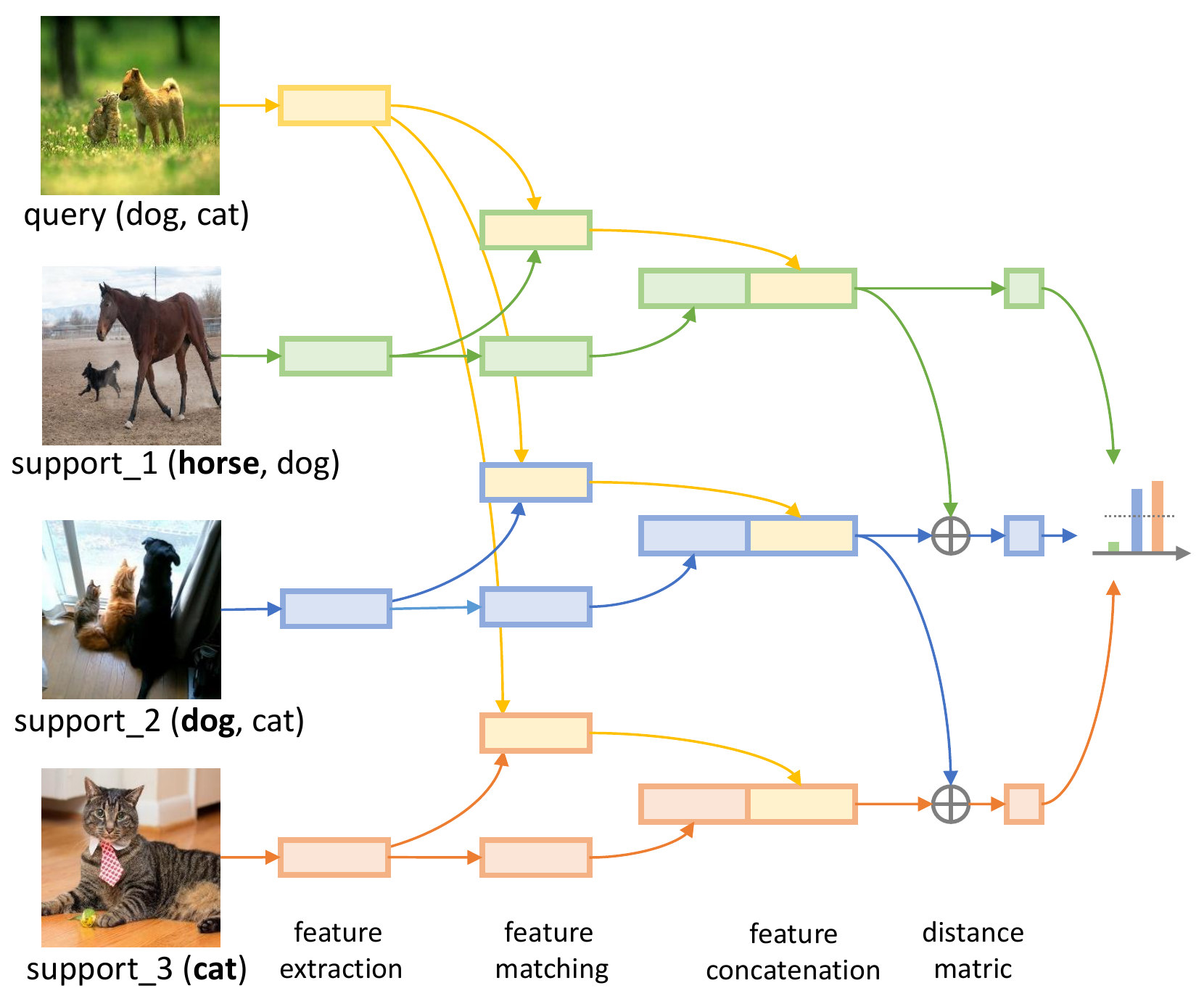}
\caption{\textbf{Illustration of CFMN for multi-label few-shot image classification.} It shows a $3$-label $3$-way $1$-shot task. The first support image is sampled as a horse image, but it also contains another interested object, \textit{i.e.}, the dog. Therefore, during measuring the distance of the query and the dog category, both the first and the second support images are considered. The concatenated features of them are averaged before the distance metric procedure.}
  \label{multi-label}
\end{figure}

We propose a multi-label extension to the traditional few-show classification problem, where each image may contain more than one interested object. In this extended setting, the mapping between images and categories is many-to-many instead of many-to-one. As shown in Fig.~\ref{multi-label}, taking $3$-way $1$-shot task for example, we first sample $3$ categories (horse, dog, cat) and sample a support image for each category form all the images that contain the object. The first support image is sampled as a horse image, but it also contains another interested object, \textit{i.e.}, the dog, and the query also belongs to more than one category. We believe that not only this setting brings up a more difficult and realistic problem to solve, but will also drive the model to learn a more generalized ability of images matching. Since the difficulty of memorization grows exponentially as the total number of categories, and the same image can become strong support but also a strong distractor under different queries. During inference, the final output is a $3$-dimensional vector. The label values higher than a particular threshold (\textit{e.g.}, 0.4) are considered positive. In Section~\ref{Experiments}, we will show that our proposed method surpasses other previous methods in this problem. 

\section{Experiments} \label{Experiments}

\begin{table}[t]\footnotesize
  \caption{\textbf{The backbone of Cascaded Feature Matching Network for different datasets.} CB: Convolution Block; CO: Concatenation Op; FCB: Fully Connected Block. The left output size is calculated based on \emph{mini}ImageNet ($84\times 84$) for example.}
  \label{architecture}
  \centering
  \begin{tabular}{>{\color{black}}l >{\color{black}} >{\color{black}}c >{\color{black}}c >{\color{black}}c >{\color{black}}c}
    \toprule
    \multirow{2}*{block name} & \multicolumn{2}{c}{{\color{black}miniImageNet \& Omniglot}} & \multicolumn{2}{c}{{\color{black}FS-COCO}} \\
    ~   &  output size  &  layers &  output size  &  layers  \\
    \midrule
    CB 1  & $41\times41\times64$ & \tabincell{c}{$3\times3$ conv, 64 filters, BN, ReLU\\ $2\times2$ maxpool, stride 2} & $56\times56\times64$ & \tabincell{c}{$7\times7$, 64, stride 2\\ $3\times3$ max pool, stride 2\\
    $\left[
  \begin{array}{c}
  3\times 3, 64 \\
  3\times 3, 64\\
  \end{array}
\right]\times 2
   $ }\\
    \midrule
    CB 2  & $19\times19\times64$ &  \tabincell{c}{$3\times3$ conv, 64 filters, BN, ReLU\\ $2\times2$ maxpool, stride 2} & $28\times28\times128$ &  \tabincell{c}{$\left[
  \begin{array}{c}
  3\times 3, 128 \\
  3\times 3, 128\\
  \end{array}
\right]\times 2
   $ }\\
    \midrule
    CB 3  & $19\times19\times64$ &  \tabincell{c}{$3\times3$ conv, 64 filters, BN, ReLU} & $14\times14\times256$ &  \tabincell{c}{$\left[
  \begin{array}{c}
  3\times 3, 256 \\
  3\times 3, 256\\
  \end{array}
\right]\times 2
   $ }  \\
    \midrule
    CB 4  & $19\times19\times64$ &  \tabincell{c}{$3\times3$ conv, 64 filters, BN, ReLU} & $7\times7\times512$ &  \tabincell{c}{$\left[
  \begin{array}{c}
  3\times 3, 512 \\
  3\times 3, 512\\
  \end{array}
\right]\times 2
   $ } \\
    \midrule
    CO   & $19\times19\times128$ &    & $7\times7\times1024$ &    \\
    \midrule
    CB 5  & $8\times8\times64$ &  \tabincell{c}{$3\times3$ conv, 64 filters, BN, ReLU\\ $2\times2$ maxpool, stride 2 } & $7\times7\times256$ &  $3\times3$, 256, stride 1\\
    \midrule
    CB 6  & $3\times3\times64$ &  \tabincell{c}{$3\times3$ conv, 64 filters, BN, ReLU\\ $2\times2$ maxpool, stride 2 }& $3\times3\times64$ &  \tabincell{c}{$3\times3$, 64, stride 1\\ $2\times2$ max pool, stride 2\\} \\
    \midrule
    FCB & $ 1$ & \tabincell{c}{$576\times8$ FC, ReLU \\ $8\times1$ FC, Sigmoid} & $ 1$ & \tabincell{c}{$576\times8$ FC \\ $8\times1$ FC, Sigmoid}  \\
    \bottomrule
  \end{tabular}
\end{table}

\subsection{Dataset}

\textbf{Omniglot}~\cite{lake2015human} was collected via Amazon’s Mechanical Turk to produce a standard benchmark for the few-shot learning task of the handwritten character recognition domain. It contains 20 examples of 1623 characters from 50 different alphabets ranging from well-established international languages which can be viewed as a transpose of the dataset MNIST. The images are resized to $28\times28$. Following \cite{vinyals2016matching,santoro2016meta}, the data set is augmented with random rotations by multiples of 90 degrees. There are $1200$ and $423$ classes for training and testing, respectively.

\textbf{\emph{mini}ImageNet} was proposed in \cite{vinyals2016matching} by sampling a subset from the well-known ImageNet dataset~\cite{russakovsky2015imagenet}. It is a large-scale and challenging few-shot image classification dataset that consists of real-world images, and it has been served as a standard benchmark for many few-shot image classification methods. \emph{mini}ImageNet contains $100$ classes, and each class has $600$ images in the size of $84\times 84$ pixels.  Because the exact train-test splits used in \cite{vinyals2016matching} were not released, we followed the splits introduced by \cite{ravi2016optimization}. In this split setting, there are $64$, $16$ and $20$ classes for training, validation, and testing, respectively. 

\textbf{FS-COCO} is the first dataset for multi-label few-shot learning proposed in this paper. It is a new split of the COCO dataset~\cite{lin2014microsoft}, which is one of the most popular datasets in multi-label classification. COCO contains $80$ classes in total. In our setting, the dataset is randomly divided into $54$, $11$, and $15$ classes for training, validation, and testing, respectively. The details of the data split can be found in the Appendix. Since the ground-truth labels of the test set are not available, we only use the samples from the training set and validation set of version $2014$ of COCO. The images are resized to $224\times224$.

\subsection{Architecture}

Most few-shot learning models utilize four convolution layers for embedding feature extractor \cite{vinyals2016matching, pmlr-v70-finn17a, sung2017learning}. For a fair comparison, we follow the same architecture for \emph{mini}ImageNet and Omniglot which is shown in Table~\ref{architecture}. Each Convolution Block contains a $3\times3$ convolution layer followed by batch normalization and a ReLU non-linearity layer. The third and the fourth Convolution Blocks do not contain the $2\times2$ max-pooling layer for providing a larger feature map to the following distance metric network. The Concatenation Operation is applied on the channel-dimension. After the Convolution Block 6, the feature is reshaped to a vector and fed to the following two Fully Connected Blocks. The final output is a single value represents the similarity of the compared images. For the multi-label few-shot classification task on FS-COCO, a structure similar to ResNet-18~\cite{he2016deep} is used which is also shown in Table~\ref{architecture}. The input size is $224\times 224$.

\subsection{Training details}

We carry out $5$-way $1$-shot and $5$-way $5$-shot image classification experiments for FS-COCO. For each episode on the $5$-way $1$-shot task, the support set is composed by sampling $1$ image from each of the $5$ classes; then we sample another $15$ samples as the query set from each of the $5$ class among the remain images for $1$-shot task; thus there are $1\times 5+15\times 5=80$ images in a \textit{episode}/mini-batch for training. As for $5$-way $5$-shot classification, there are $5$ images for each class in the support and query set, respectively. Following~\cite{snell2017prototypical}, the model is trained on $20$-way and $30$-way $15$ queries per training episode for \emph{mini}ImageNet. Beside $5$-way $1$-shot and $5$-way $5$-shot, $20$-way for $1$-shot and $5$-shot image classification experiments are also evaluated on Omniglot. There are $19$ and $15$ images for each class in the query set for $1$-shot and $5$-shot, respectively. 

Our few-shot image classification network is trained on the training set and validated on the validation set. The model that obtains the best performance on the validation set is selected. The selected model is evaluated on the testing set to obtain the final results. The mean square error (MSE) loss is used to train our model.

We implement the proposed network using PyTorch~\cite{paszke2017automatic}. The optimizer is Adam\cite{kinga2015method}; the learning rate decreases by $0.1$ to the original one if the validation accuracy does not increase during the last $15,000$ \textit{episode}. Besides, the current best model will be reloaded and trained with the updated learning rate. The training procedure is early stopped if the validation accuracy does not increase during the last $50,000$ \textit{episode}.

\subsection{Testing details}

In testing and validation, there are $600$ episodes for datasets MS-COCO and \emph{mini}ImageNet. In every episode, $1$ and $5$ support images per class are sampled for the $1$-shot setting and the $5$-shot setting, respectively. Then $15$ images for each class are taken as the queries. Thus, we have $45,000 = 600 \times 15 \times 5$ classification results. The mean and confidence intervals of the classification accuracy of the $45,000$ testings are recorded. For dataset Omniglot, there are $1000$ testing episodes. In every episode, $19$ and $15$ query images per class are sampled for the $1$-shot and the $5$-shot setting, respectively.

To avoid the randomness of the episode sampling effects, we perform the above testing procedure for $10$ times. The mean of the accuracy and confidence intervals over all the $10$ times are reported in this paper.

\subsection{Results} \label{result}

\begin{table*}[t]\footnotesize
  \caption{\textbf{Few-shot images classification accuracies on Omniglot.} `-': not reported. The best results are bold. The Cascaded Feature Matching Network (CFMN) obtains the state-of-the-art or comparable performance on all settings. Some accuracy results are reported with $95\%$ confidence intervals.}
  \label{results_omniglot}
  \centering
  \begin{tabular}{lccccc}
    \toprule
    Methods    & Ref & $5$-way $1$-shot     & $5$-way $5$-shot    & $20$-way $1$-shot     & $20$-way $5$-shot\\
    \midrule
    MANN \cite{santoro2016meta} & ICML'16 & $82.8\%$ & $94.9\%$ & - & - \\
    Matching Network \cite{vinyals2016matching} & NIPS'16 & $98.1\%$ & $98.9\%$ & $93.8\%$ & $98.5\%$ \\
    Neural Statistician \cite{edwards2016towards} & ICLR'17 & $98.1\%$ & $99.5\%$ & $93.2\%$ & $98.1\%$ \\
    ConvNet with Memory Module \cite{kaiser2017learning} & ICLR'17 & $98.4\%$ & $99.6\%$ & $95.0\%$ & $98.6\%$ \\
    Meta Network \cite{pmlr-v70-munkhdalai17a} & ICML'17 & $99.0\%$ & - & $97.0\%$ & - \\
    Prototypical Network \cite{snell2017prototypical} & NIPS'17 & $98.8\%$ & $99.7\%$ & $96.0\%$ & $98.9\%$ \\
    MAML \cite{pmlr-v70-finn17a} & ICML'17 & {\color{black} $98.7\% \pm 0.4\%$} &{\color{black} $\mathbf{99.9\%} \pm 0.1\%$ }& {\color{black}$95.8\% \pm 0.3\%$ }& {\color{black}$98.9\% \pm 0.2\%$} \\
    Relation Network \cite{sung2017learning} & CVPR'18 & {\color{black}$99.6\% \pm 0.2\%$} & {\color{black}$\mathbf{99.8\% \pm 0.1\%}$} & {\color{black}$97.6\% \pm 0.2\%$} & {\color{black}$99.1\% \pm 0.1\%$} \\
    CFMN (Ours) &  & {\color{black}$\mathbf{99.7\% \pm 0.2\%}$} & {\color{black}$\mathbf{99.8\% \pm 0.1\%}$} & {\color{black}$\mathbf{98.0\% \pm 0.2\%}$} & {\color{black}$\mathbf{99.2\% \pm 0.1\%}$} \\
    \bottomrule
  \end{tabular}
\end{table*}

\begin{table*}[ht]\footnotesize
  \caption{\textbf{Few-shot images classification accuracies on \emph{mini}ImageNet.} `-': not reported. The best results are bold. The Cascaded Feature Matching Network (CFMN) obtains the state-of-the-art performance on $5$-way $1$-shot and competitive results on $5$-way $5$-shot. All accuracy results are reported with $95\%$ confidence intervals.}
  \label{results_mini}
  \centering
  \begin{tabular}{lccc}
    \toprule
    Methods  & Ref    & $5$-way $1$-shot     & $5$-way $5$-shot \\
    \midrule
    Matching Network \cite{vinyals2016matching} & NIPS'16  & $43.56\% \pm 0.84\%$ & $55.31\% \pm 0.73\%$ \\
    Meta Network \cite{pmlr-v70-munkhdalai17a}  & ICML'17   & $49.21\% \pm 0.96\%$ &  - \\
    Meta-Learn LSTM \cite{ravi2016optimization}  & ICLR'17   & $43.44\% \pm 0.77\%$ & $60.60\% \pm 0.71\%$ \\
    MAML \cite{pmlr-v70-finn17a} & ICML'17  & $48.70\% \pm 1.84\%$ & $63.11\% \pm 0.92\%$ \\
    Prototypical Network \cite{snell2017prototypical} & NIPS'17 & $49.42\% \pm 0.78\%$ & $68.20\% \pm 0.66\%$ \\
    Relation Network \cite{sung2017learning} & CVPR'18  & $50.44\% \pm 0.82\%$ & $65.32\% \pm 0.70\%$ \\
    CFMN (Ours) & & $\mathbf{52.98\% \pm 0.84\%}$ & $\mathbf{68.33\% \pm 0.70\% }$ \\
    \bottomrule
  \end{tabular}
\end{table*}

\begin{table*}[th]\footnotesize
  \caption{\textbf{Multi-label few-shot images classification accuracies on FS-COCO.} The best results are bold. CFMN obtains the best performance.}
  \label{results-coco}
  \centering
  \begin{tabular}{lcccccc}
    \toprule
    \multirow{2}{*}{Model} & \multicolumn{3}{c}{$5$-way $1$-shot} & \multicolumn{3}{c}{$5$-way $5$-shot}\\
    \cline{2-7}
    ~ & Precision & Recall & F1 & Precision & Recall & F1\\
    \midrule
    Prototypical Network \cite{snell2017prototypical} & $32.78\%$ & $45.96\%$ & $38.06\%$ & $44.42\%$ & $61.10\%$ & $51.22\%$ \\
    Relation Network \cite{sung2017learning} & $34.37\%$ & $47.21\%$ & $39.52\%$ & $43.61\%$ & $63.34\%$ & $51.43\%$\\
    CFMN (Ours) & $\mathbf{37.61}\%$ & $\mathbf{53.90}\%$ & $\mathbf{44.14}\%$ & $\mathbf{45.71}\%$ & $\mathbf{64.46}\%$ & $\mathbf{53.25}\%$\\
    \bottomrule
  \end{tabular}
\end{table*}

\begin{table}[th]\footnotesize
  \caption{\textbf{Impact of weight factor of matched feature. } All the results are evaluated on \emph{mini}ImageNet for $5$-way $1$-shot task. The best results are bold.}
  \label{weight}
  \centering
  \begin{tabular}{lc}
    \toprule
    weight factor     & accuracy   \\
    \midrule
    CFMN with $\lambda=0.00$ & $50.89\% $   \\
    CFMN with $\lambda=0.25$ & $52.02\% $  \\
    CFMN with $\lambda=0.50$ & $\mathbf{52.98\% }$  \\
    CFMN with $\lambda=0.75$ & $50.28\% $  \\
    CFMN with $\lambda=1.00$ & $45.59\% $ \\
    \bottomrule
  \end{tabular}
\end{table}

\begin{table}[t]\footnotesize
  \caption{\textbf{Impact of the details of the feature matching block. } All the results are evaluated on \emph{mini}ImageNet for $5$-way $1$-shot task.}
  \label{table:ablation}
  \centering
  \begin{tabular}{lclc}
    \toprule
    Model & accuracy & Model & accuracy \\
    \midrule
    $C_m=4$ & $51.63\% $ & $C_m=64$ & $52.98\% $ \\
    $C_m=8$ & $52.14\% $ & $C_m=128$ & $52.49\% $ \\
    $C_m=16$ & $52.52\% $ & w/o softmax & $49.93\% $ \\
    $C_m=32$ & $52.93\% $ & w/o transformation & $52.46\% $ \\
    \bottomrule
  \end{tabular}
\end{table}

\noindent
\textbf{Results on Omniglot and \emph{mini}ImageNet}

Table~\ref{results_omniglot} and~\ref{results_mini} illustrate the performance of our method against the current state-of-the-art on Omniglot and \emph{mini}ImageNet, respectively. All accuracy results are reported with $95\%$ confidence intervals. The best performing results are bold. It can be observed that our CFMN obtains better performance on both the two classical benchmarks than the state-of-the-art models, such as Relation Network, MAML, Prototypical Network, Meta Network. 

\begin{figure*}[t]
  \centering
  \includegraphics[width=1\linewidth]{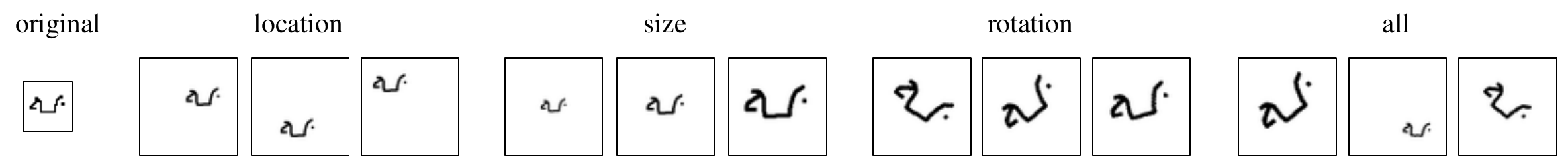}
\caption{\textbf{Samples of four harder variations on Omniglot.} Original: Image size is $28\times 28$. The characters are always in the center. Location: Images of the original set are randomly put in a $56\times 56$ white background. Size: Characters are randomly resized to $[20,55]$, and put in the center of the $56\times 56$ white background. Rotation: Characters are resized to $50$, and randomly rotated $[-45,45]$ degrees, and put in the center of the $56\times 56$ white background. All: Characters are randomly resized to $[20,55]$, and randomly rotated $[-45,45]$ degrees, and randomly put in the $56\times 56$ white background.}
  \label{omniglot}
\end{figure*}

\vspace{2mm}
\noindent
\textbf{Multi-label few-shot learning results on FS-COCO}

As shown in Table~\ref{results-coco}, precision, recall, and F1-measure are employed to evaluate the models. Labels with confidence higher than $0.4$ are considered positive. These measures do not require a fixed number of labels per image. Our model outperforms the existing methods by a sizable margin.

\vspace{2mm}
\noindent
\textbf{Impact of weight factor of matched feature}

As defined in Sect.~\ref{FMB}, $\lambda$ represents the ratio of the matched feature and the original feature. $\lambda=0.0$ means only using the original feature while $\lambda=1.0$ means that only the matched feature is taken into account. We evaluated our model with several standard values for $\lambda$. Referring to the results shown in Table~\ref{weight}, it can be found that the model cannot reach the best performance with whether the original feature alone or the matched feature alone. When $\lambda=1$, the network only takes the matched information into consideration. But shallow layers only get some low-level vision information like the color, shape, and edge. Although feature matching is really helpful, an appropriate combination of the matched feature and the original feature is necessary. Making an analogy with how our human beings recognize the similarity of two images, we would not only compare the details of them but also conclude by the visual context of the whole image. The combination by the ratio $\lambda$ behaves in the same way.

\vspace{2mm}
\noindent
\textbf{Impact of details of the feature matching block}

Table~\ref{table:ablation} shows the impact of the reduction dim $C_m$, the softmax axis and space transformation operation. It can be seen that the accuracy does not just simply improve as $C_m$ increases. An appropriate setting for $C_m$ can get better performance, at the same time reduce the computation. The row-wise softmax and space transformation both directly improve accuracy. But obviously, the row-wise softmax is more important to the results.

\vspace{2mm}
\noindent
\textbf{Impact of the cascaded structure} 

\begin{table}[t]\footnotesize
  \caption{\textbf{Impact of the cascaded structure. } All the results are evaluated on \emph{mini}ImageNet for $5$-way $1$-shot task. The best results are bold.}
  \label{cascaded}
  \centering
  \begin{tabular}{lclc}
    \toprule
    layers     & accuracy  & layers     & accuracy \\
    \midrule
    CB $1$ & $50.47\% $ & CB $3,4$ & $52.34\% $\\
    CB $2$ & $51.11\% $ & CB $2,3,4$ & $\mathbf{52.98\% }$\\
    CB $3$ & $51.63\% $ & CB $1,2,3,4$ & $50.17\%$ \\
    CB $4$ & $51.92\% $ \\
    \bottomrule
  \end{tabular}
\end{table}

As defined in Sect.~\ref{CFMN}, we take a cascaded structure for combining the matched information from different representation levels to reach a more accurate and robust performance. In order to illustrate the necessity and effectiveness of this structure, we applied the different numbers of feature matching blocks in different positions at the backbone. For example, Convolution Block $1,2,3,4$ means that there are four feature matching blocks after the first fourth Convolution Blocks, respectively. From the results in Table~\ref{cascaded}, we can see that if taking only one feature matching block, deeper layers are better than the shallow one. Table~\ref{cascaded} also shows that the cascaded structure is much better than only a single one feature matching block. However, the exception is that the feature after the first Convolution Block is unsuitable for the matching block. Because the feature merely contains pixel information. Applying feature matching block here will make the model focus too much on the low-level feature which has detrimental effects on the performance.

\subsection{How does CFMN work} \label{harder}

\noindent
\textbf{The effectiveness of spatial feature matching} 

\begin{table*}[t]
  \caption{\textbf{Results of four harder settings on Omniglot on $10$-way $1$-shot task.} The best results are bold. Our CFMN always reaches the best performances. It can greatly reduce the influence caused by the differences in the size, location, rotation and even the combination of them. }
  \label{effectiveness}
  \centering
  \begin{tabular}{lccccc}
    \toprule
    weight factor  &  original &  size  &  location   &   rotation  &  all   \\
    \midrule
    Prototypical Network \cite{snell2017prototypical} & {\color{black}$98.02\% $} & $95.75\% $ & $94.34\% $ & $93.67\% $ & $88.93\%$  \\
    Relation Network \cite{sung2017learning} & {\color{black}$99.18\% $} & $98.95\% $ & $97.64\% $ & $96.94\% $ & $94.95\% $  \\
    CFMN (Ours) & {\color{black}$\mathbf{99.23\% }$} & $\mathbf{98.99\% }$ & $\mathbf{99.05\% }$ & $\mathbf{98.42\% }$ & $\mathbf{97.89\%}$  \\
    \bottomrule
  \end{tabular}
\end{table*}

In order to further check the effectiveness of the feature matching block, we design four harder variations (query and support images are highly variant in \emph{location}-variation, \emph{size}-variation, \emph{rotation}-variation and \emph{all}-variation). As shown in Fig.~\ref{omniglot}, the image size of all the four harder variations is $56\times 56$. Each image in the Omniglot is used to create 10 different images. In the \emph{location}-variation, we randomly place the handwritten character on a white background. For the \emph{size}-variation, we randomly resize each character by the size range in $[20,55]$ and put it in the center of a white background. Analogously, each image is randomly rotated by $-45$ to $45$ degrees for the \emph{rotation}-variation. The rotated images are also put in the center of a white background. As for \emph{all}-variation, as the name implies, it combines all of the former operations for each image, which is more difficult than the other.

We evaluate our CFMN on all the four harder variations and compare it with two existing methods. We can see from the results shown in Table~\ref{effectiveness} that CFMN consistently outperforms the other works, especially on the \emph{all}-variation. {\color{black}The results on original Omniglot data are similar to each other.} But the performances of Matching Network and Prototypical Network severely decrease when dealing with harder visual differences. It illustrates that our proposed model can overcome the obstacles from the object variations in the size, rotation, location, and even the combination of them.

\vspace{2mm}

\noindent
\textbf{Visualization} 

\begin{figure*}[t]
  \centering
  \includegraphics[width=1\linewidth]{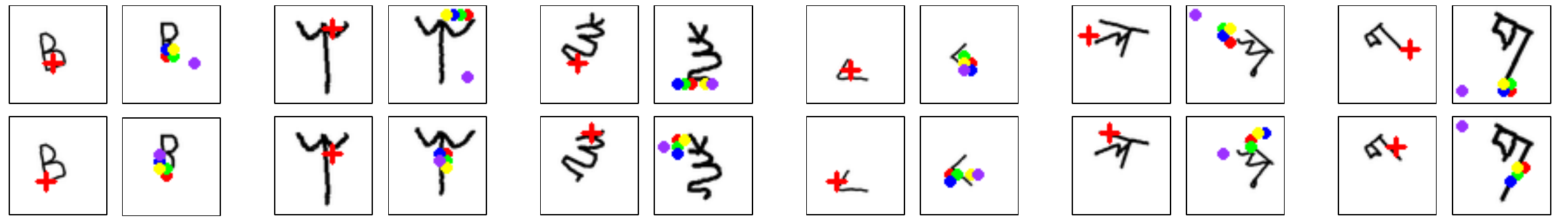}
  \caption{\textbf{Visualization of feature matching on the \emph{all}-variation of Omniglot defined in Sect~\ref{harder}.} Two adjacent images form a group. The left one is the query. The red cross in it is an image position which is matched with all positions of the right support image. The colours, in turn, the \textcolor{red}{red}, \textcolor{green}{green}, \textcolor{black}{black}, \textcolor{yellow}{yellow} and \textcolor{mypurple}{purple} point the positions which have the top five highest correlation responses.}
  \label{visualization_omniglot}
\end{figure*}

\begin{figure*}[t]
  \centering
  \includegraphics[width=1\linewidth]{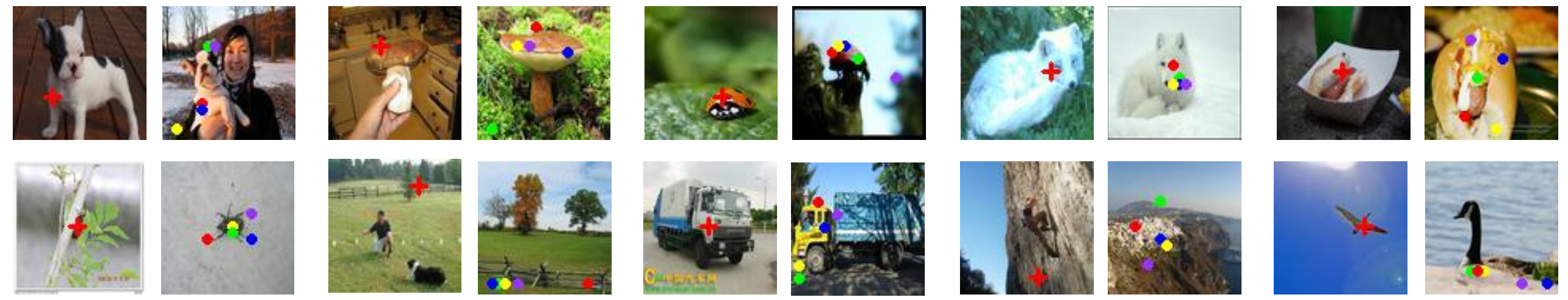}
  \caption{\textbf{Visualization of feature matching on \emph{mini}ImageNet.} The meaning of the red cross and colored dot is the same as Fig.~\ref{visualization_omniglot}. Although the interested objects of each class may be different in the size, location, style and so on, they are associated together by our matching operation.}
  \label{visualization}
\end{figure*}

\begin{figure*}[t]
  \centering
  \includegraphics[scale=0.38]{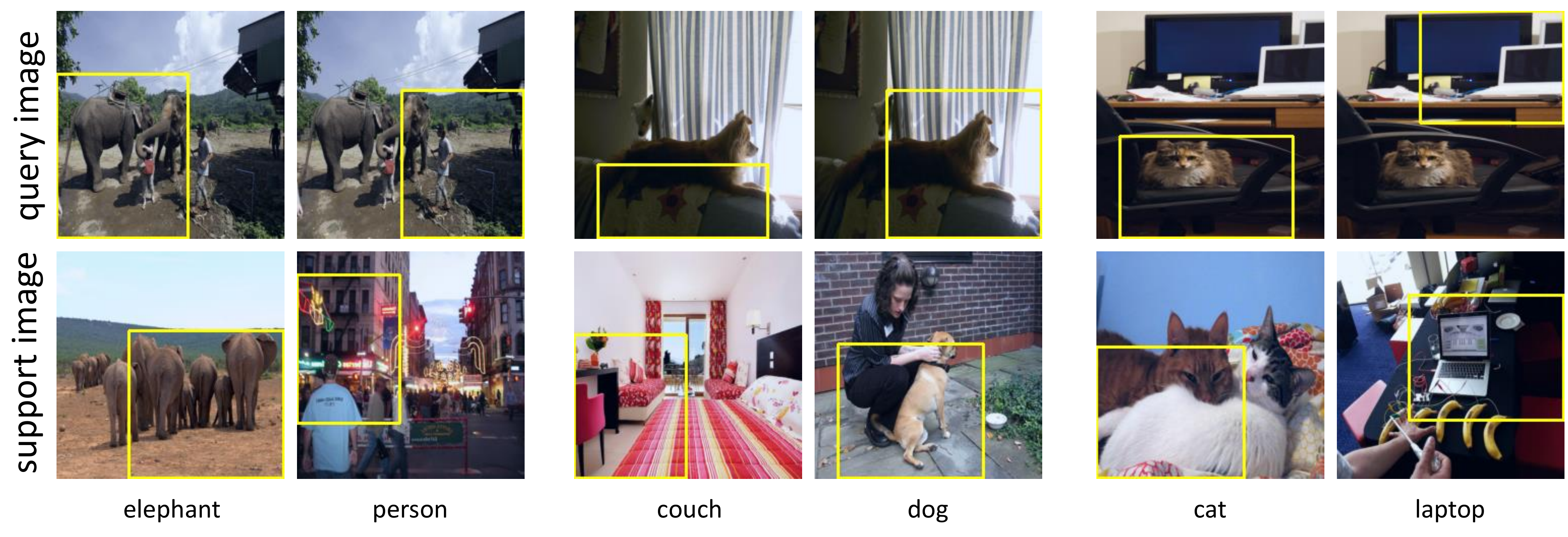}
  \caption{\textbf{Visualization of feature matching on FS-COCO.} The yellow rectangular boxes indicate the receptive fields of the features that get the highest correlation responses in the last Feature Matching Block. Two images aligned vertically is a group. }
  \label{visualization_coco}
\end{figure*}

To provide a more intuitive view of how our proposed method works, we visualize the feature matching operation in Fig.~\ref{visualization_omniglot}, Fig.~\ref{visualization} and Fig.~\ref{visualization_coco}. Two images from the same class form a group in Fig.~\ref{visualization_omniglot} and Fig.~\ref{visualization}. The left is the query; the right is the support image. {\color{black}The visualization is based on the spatial attention map in the last feature matching block. It stands for the performance of all of the three matching blocks because a matched feature is also the input of the next matching block. The feature has been matched three times after all of three matching blocks. The position represented by a red cross in the query is matched with all the right positions. By comparing the values in the spatial attention map, we point positions which have the top five highest correlation responses by different colors.} It can be seen from figures that although the compared characters are different in the size, location, rotation, and so on, the corresponding strokes are associated together by our matching operation. 

Since a deeper network is used for FS-COCO, receptive fields of the features in the last Feature Matching Block is larger than them in \emph{mini}ImageNet and Omniglot. Therefore, the receptive field is indicated by the rectangular box in Fig.~\ref{visualization_coco}. We can find that when the same query image matched with different support images, the associated parts can get higher responses in the spatial attention map, which benefits a lot in the multi-label few-shot setting. 

\section{Conclusion} \label{Conclusion}

In this paper, we proposed the Cascaded Feature Matching Network (CFMN), which is a simple and effective method for few-shot image recognition. Our motivation is based on the observation that the interested object in compared images from the real world usually differs significantly in the size, location, style, \textit{etc}. Our feature matching block can overcome those barriers and associate the corresponding parts together. The features with high correlation responses are paid more attention, while the opposite will be naturally ignored. Three feature matching blocks are applied there to construct a cascaded structure that combines the matching information from different representation levels. The extensive experiments on few-shot and multi-label few-shot classification on three standard datasets demonstrate the effectiveness of our proposed method. 

\section*{Appendix}
{\color{black}\textbf{Data split for FS-COCO}}

{\color{black}\textbf{Training set:} toilet, teddy bear, bicycle, skis, tennis racket, snowboard, carrot, zebra, keyboard, scissors, chair, couch, boat, sheep, donut, tv, backpack, bowl, microwave, bench, book, elephant, orange, tie, bird, knife, pizza, fork, hair drier, frisbee, bottle, bus, bear, toothbrush, spoon, giraffe, sink, cell phone, refrigerator, remote, surfboard, cow, dining table, hot dog, baseball bat, skateboard, banana, person, train, truck, parking meter, suitcase, cake, traffic light.}

{\color{black}\textbf{Validation set:} sandwich, kite, cup, stop sign, toaster, dog, bed, vase, motorcycle, handbag, mouse.}

{\color{black}\textbf{Testing set:} laptop, horse, umbrella, apple, clock, car, broccoli, sports ball, cat, baseball glove, oven, potted plant, wine glass, airplane, fire hydrant.}

\Acknowledgements{This work was supported by NSFC (No. 61876212, No. 61733007 and No. 61572207) and HUST-Horizon Computer Vision Research Center.}

\bibliographystyle{plain}
\bibliography{cfmn}






\end{document}